\documentclass[10pt,twocolumn,letterpaper]{article}

\usepackage{cvpr}
\usepackage{times}
\usepackage{epsfig}
\usepackage{graphicx}
\usepackage{amsmath}
\usepackage{amssymb}
\usepackage{bm}
\usepackage{subcaption}
\usepackage{epstopdf}
\usepackage{subcaption}
\usepackage{enumitem}

\cvprfinalcopy 

\def\cvprPaperID{2136} 

\begin{document}

\title{Robust 3D Hand Pose Estimation in Single Depth Images: \\
	from Single-View CNN to Multi-View CNNs}

\author{Liuhao Ge, Hui Liang, Junsong Yuan and Daniel Thalmann\\
Institute for Media Innovation\\
Nanyang Technological University, Singapore\\
{\tt\small \{ge0001ao, hliang1\}@e.ntu.edu.sg, \{jsyuan, danielthalmann\}@ntu.edu.sg}
}

\maketitle

\begin{abstract}
Articulated hand pose estimation plays an \mbox{important} role in human-computer interaction. Despite the \mbox{recent} progress, the accuracy of existing methods is still not satisfactory, partially due to the difficulty of embedded high-dimensional and non-linear regression problem. \mbox{Different} from the existing discriminative methods that regress for the hand pose with a single depth image, we propose to first project the query depth image onto three orthogonal planes and utilize these multi-view projections to regress for 2D heat-maps which estimate the joint positions on each plane. These multi-view heat-maps are then fused to produce final 3D hand pose estimation with learned pose priors. Experiments show that the proposed method largely outperforms state-of-the-art on a challenging dataset. Moreover, a cross-dataset experiment also demonstrates the good generalization ability of the proposed method.
\end{abstract}

\section{Introduction}
\label{Introduction}
The problem of 3D hand pose estimation has aroused a lot of attention in computer vision community for long, as it plays a significant role in human-computer interaction such as virtual/augmented reality applications. Despite the recent progress in this field~\cite{oberweger2015training,ren2013robust,sun2015cascaded,tagliasacchi2015robust,tompson14tog}, robust and accurate hand pose estimation remains a challenging task. Due to large pose variations and high dimension of hand motion, it is generally difficult to build an efficient mapping from image features to articulated hand pose parameters.

Data-driven methods for hand pose estimation train discriminative models, such as isometric self-organizing map~\cite{PTREF29}, random forests ~\cite{keskin2012hand,sun2015cascaded,tang2014latent,tang2013real} and convolutional neural networks (CNNs)~\cite{tompson14tog}, to map image features to hand pose parameters. With the availability of large annotated hand pose datasets~\cite{sun2015cascaded,tang2014latent,tompson14tog}, data-driven approaches become more advantageous as they do not require complex model calibration and are robust to poor initialization. 

\begin{figure}[t]
	\begin{center}
		\includegraphics[width=1.0\linewidth]{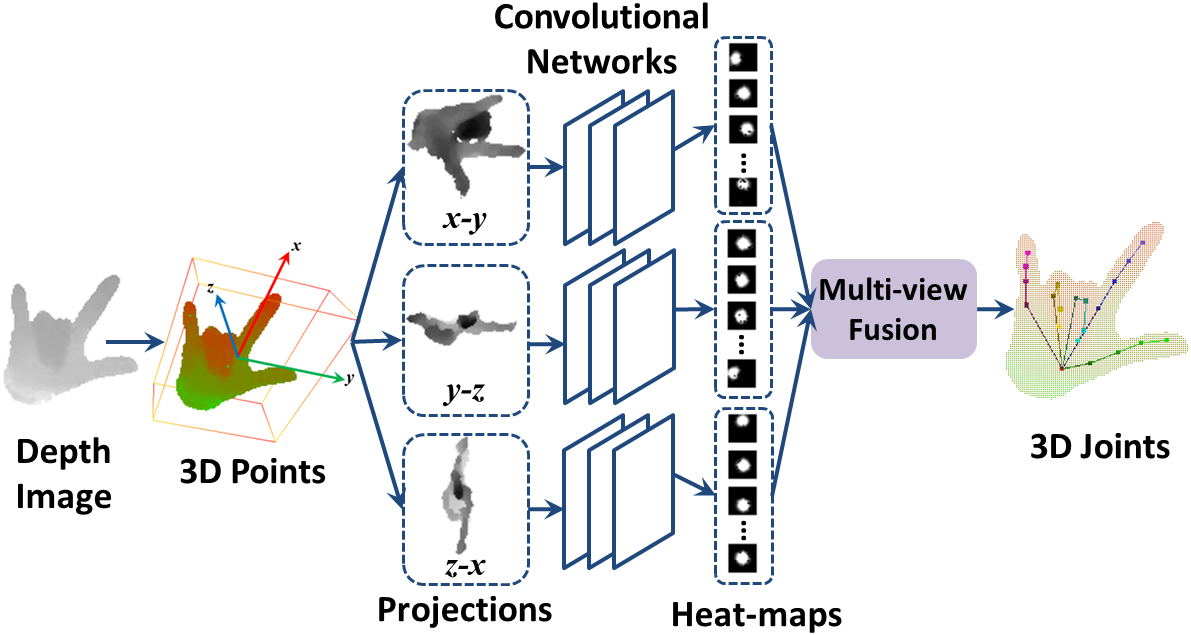}
	\end{center}
	\caption{Overview of our proposed multi-view regression framework. We generate heat-maps for three views by projecting 3D points onto three orthogonal planes. Three \mbox{CNNs} are trained in parallel to map each view's projected image to its corresponding heat-maps, which are then fused together to estimate 3D hand joint locations.}
	\label{fig:Fig_Framework}
\end{figure}

We focus on CNN-based data-driven methods in this paper. CNNs have been applied in body and hand pose estimation~\cite{tompson2015efficient,tompson14tog,toshev2014deeppose} and have shown to be effective. The main difficulty of CNN-based methods for hand pose estimation lies in accurate 3D hand pose regression. Direct mapping from input image to 3D locations is highly non-linear with high learning complexity and low generalization ability of the networks~\cite{tompson2015efficient}. One alternative way is to map input image to a set of heat-maps which represent the probability distributions of joint positions in the image and recover the 3D joint locations from the depth image with model fitting~\cite{tompson14tog}. However, in this method, the heat-map only provides 2D information of the hand joint and the depth information is not fully utilized.

In this work, we propose a novel 3D regression method using multi-view CNNs that can better exploit depth cues to recover fully 3D information of hand joints without model fitting, as illustrated in Fig.~\ref{fig:Fig_Framework}. Specifically, the point cloud of an input depth image is first projected onto three orthogonal planes, and each projected image is then fed into a separate CNN to generate a set of heat-maps for hand joints following similar pipeline in~\cite{tompson14tog}. As the heat-map in each view encodes the 2D distribution of a joint on the projection plane, their combination in three views thus contains the location distribution of the joint in 3D space. By fusing heat-maps of three views with pre-learned hand pose priors, we can finally obtain the 3D joint locations and alleviate ambiguous estimations at the same time.

\begin{figure}[t]
	\centering
	\begin{subfigure}[t]{0.9in}
		\includegraphics[width=0.9in]{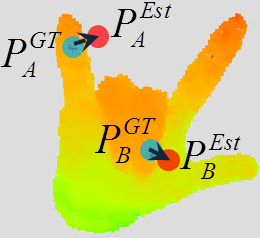}
		\caption{}
		\label{fig:problem_depth}
	\end{subfigure}
	~
	\begin{subfigure}[t]{2.28in}
		\includegraphics[width=2.28in]{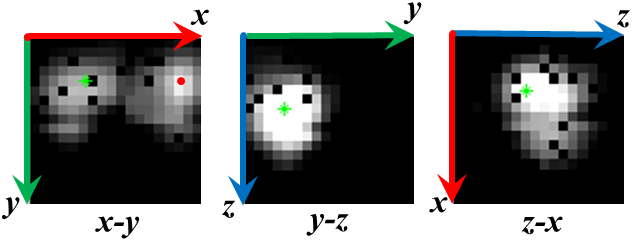}
		\caption{}
		\label{fig:problem_heatmap}
	\end{subfigure}
	\caption{(a) Illustration of joint estimation in single view. Blue points are true locations, and red points are estimated locations. The little finger tip is misestimated on the background and the middle finger tip is misestimated on the hand palm. (b) Illustration of ambiguous estimation. Despite the heat-map of \textit{x}-\textit{y} view contains two hotspots which are hard to choose, from the heat-map of \textit{z}-\textit{x} view, it is easy to find that the \textit{x} value is small with high confidence. Thus, the left hotspot in \textit{x}-\textit{y} view's heat-map is true.}
	\label{fig:Fig_Intro1}
\end{figure}

Compared to the method of single view CNN in~\cite{tompson14tog}, our proposed method of multi-view CNNs has the following advantages:
\begin{itemize}[leftmargin=*]\itemsep0pt
	\item In the single view CNN, the depth of a hand joint is taken as the corresponding depth value at the estimated 2D position, which may result in large depth estimation errors even if the estimated 2D position is only slightly deviated from the true joint position, as shown in Fig.~\ref{fig:problem_depth}. In contrast, our proposed multi-view CNNs generate heat-maps for front, side and top views simultaneously, from which the 3D locations of hand joints can be estimated more robustly.
	\item In case of ambiguous estimations, the single view \mbox{CNN} cannot well differentiate among multiple hotspots in the heat-map, in which only one could correspond to the true joint, as shown in Fig.~\ref{fig:problem_heatmap} (\textit{x}-\textit{y} view). With the proposed multi-view CNNs, the heat-maps from other two views can help to eliminate the ambiguity, such as that in Fig.~\ref{fig:problem_heatmap}.
	\item Different from~\cite{tompson14tog} that still relies on a pre-defined hand model to obtain the final estimation, our proposed approach embeds hand pose constraints learned from training samples in an implicit way, which allows to enforce hand motion constraints without manually defining hand size parameters.
\end{itemize}

Comprehensive experiments validate the superior performance of the proposed method compared to state-of-the-art methods on public datasets~\cite{sun2015cascaded}, with runtime speed of over 70fps. In addition, our proposed multi-view regression method can achieve relatively high accuracy in cross-dataset experiments~\cite{qian2014realtime, sun2015cascaded}.

\section{Literature Review}
Vision-based hand pose estimation has been extensively studied in literature over many years. The most common hand pose estimation techniques can be classified into model-driven approaches and data-driven approaches~\cite{supancic2015depth}. Model-driven methods usually find the optimal hand pose parameters via fitting a deformable 3D hand model to input image observations. Such methods have demonstrated to be quite effective, especially with the depth cameras~\cite{oikonomidis2011efficient,qian2014realtime,PTREF27,tagliasacchi2015robust}. However, there are some shortcomings for the model-driven methods. For instance, they usually need to explicitly define the anatomical size and hand motion constraints of the hand to match to the input image. Also, due to the high dimensional of hand pose parameters, they can be sensitive to initialization for the iterative model-fitting procedure to converge to the optimal pose.

In contrast, the data-driven methods do not need the explicit specification of the hand size and motion constraints. Rather, such information is automatically encoded in the training data. Therefore, many recent methods are built upon such a scheme~\cite{liang2016barehanded,liang2015parsing,liang2015AR,sun2015cascaded,tang2013real,xu2013efficient}. Among them, the random forest and its variants have proved to be reasonably accurate and fast. In~\cite{xu2013efficient}, the authors propose to use the random forest to directly regress for the hand joint angles from depth images, in which a set of spatial-voting pixels cast their votes for hand pose independently and their votes are clustered into a set of candidates. The optimal one is determined by a verification stage with a hand model. A similar method is presented in~\cite{tang2013real}, which further adopts transfer learning to make up for the inconsistence between synthesis and real-world data. As the estimations from random forest can be ambiguous for complex hand postures, pre-learned hand pose priors are sometimes utilized to better fuse independently predicted hand joint distributions~\cite{PTREF38,liang2014resolving}. In~\cite{sun2015cascaded}, the cascaded pose regression algorithm~\cite{dollar2010cascaded} is adapted to the problem of hand pose estimation. Particularly, the authors propose to first predict the root joints of the hand skeleton, based on which the rest joints are updated. In this way the hand pose constraints can be well preserved during pose regression. 

Very recently, convolutional neural networks have shown to be effective in articulated pose estimation. In~\cite{toshev2014deeppose}, they are tuned to regress for the 2D human poses by directly minimizing the pose estimation error on the training data. The results have shown to outperform the traditional methods largely. However, it takes more than twenty days to train the network and the dataset only contains several thousand images. Considering the relatively small size of the dataset used in \cite{toshev2014deeppose}, it can be difficult to use it on larger datasets such as~\cite{sun2015cascaded,tang2014latent}, which consist of more than 70K images. Also, it is reported in~\cite{jain2013learning,tompson2015efficient} that such direct mapping with CNNs from image features to continuous 2D/3D locations is of high nonlinearity and complexity as well as low generalization ability, which renders it difficult to train CNNs in such a manner. To this end, in their work on body pose estimation~\cite{tompson2015efficient, tompson2014joint}, the CNNs are used to predict the heat-maps of joint positions instead of the original articulated pose parameters, and on each heat-map the intensity of a pixel indicates the likelihood for a joint occurring there. During training, the regression error is instead defined as the L2-norm of the difference between the estimated heat-map and the ground truth heat-map. In this way, the network can be trained efficiently and they achieve state-of-the-art performances. Similarly, such a framework has also been applied in 3D hand pose estimation~\cite{tompson14tog}. However, the heat-map only provides 2D information of the hand joint and the depth information is not fully utilized. To address this issue, a model-based verification stage is adopted to estimate the 3D hand pose based on the estimated heat-maps and the input depth image~\cite{tompson14tog}. Such heat-map based approaches are interesting as heat-maps can reflect the probability distribution of 3D hand joints in the projection plane. Inspired by such methods, we generate heat-maps of multiple views and fuse them together to estimate the probability distribution of hand joints in 3D space.

\section{Methodology}
\label{Methodology}

The task of the hand pose estimation can be regarded as the extraction of the 3D hand joint locations from the depth image. Specifically, the input of this task is a cropped depth image only containing a human hand with some gesture and the outputs are ${K}$ 3D hand joint locations which represent the hand pose. Let the ${K}$ objective hand joint locations be ${\bm{\Phi } = \left\{ {{{\bm{\phi}} _k}} \right\}_{k = 1}^K \in \bm{\Lambda }}$, here ${\bm{\Lambda}}$ is the ${3\times{K}}$ dimensional hand joint space, and in this work ${K=21}$. The 21 objective hand joint locations are the wrist center, the five metacarpophalangeal joints, the five proximal interphalangeal joints, the five distal interphalangeal joints and the five finger tips.

Following the discussion in Section~\ref{Introduction}, we propose to infer 3D hand joint locations \bm{$\Phi$} based on the projected images on three orthogonal planes. Let the three projected images be $\textit{I}_{xy}$, $\textit{I}_{yz}$ and $\textit{I}_{zx}$, which are obtained by projecting 3D points from the depth image onto \textit{x}-\textit{y}, \textit{y}-\textit{z} and \textit{z}-\textit{x} planes in the projection coordinate system, respectively. Thus, the query depth image $\textit{I}_{D}$ is transformed to the three \mbox{projections} $\textit{I}_{xy}$, $\textit{I}_{yz}$ and $\textit{I}_{zx}$, which will be used as the inputs to infer 3D hand joint locations in our following derivations.

We estimate the hand joint locations \bm{$\Phi$} by applying the MAP (maximum a posterior) estimator on the basis of projections ${\textit{I}_{xy}}$, ${\textit{I}_{yz}}$ and ${\textit{I}_{zx}}$, which can be viewed as the observations of the 3D hand pose. Given ${\left(I_D, \bm{\Phi}\right)}$, we assume that the three projections ${\textit{I}_{\textit{xy}}}$, ${\textit{I}_{\textit{yz}}}$ and ${\textit{I}_{\textit{zx}}}$ are independent, conditioned on the joint locations ${\bm{\Phi}}$~\cite{ando2007two, xu2009multimodal}. Under this assumption and the assumption of equal a priori probability ${P\left( \bm{\Phi } \right)}$, the posterior probability of joint locations can be formulated as the product of the individual estimations from all the three views. The problem to find the optimal hand joint locations ${\bm{\Phi }^ * }$ is thus formulated as follows:
\begin{equation}
\begin{aligned}
{\bm{\Phi }^ * } = & \mathop {\arg \max }\limits_{\bm{\Phi }} P\left( {\left. \bm{\Phi } \right|{I_{xy}},{I_{yz}},{I_{zx}}} \right)\\
= & \mathop {\arg \max }\limits_{\bm{\Phi }} P\left( {\left. {{I_{xy}},{I_{yz}},{I_{zx}}} \right|\bm{\Phi }} \right)\\
= & \mathop {\arg \max }\limits_{\bm{\Phi }} P\left( {\left. {{I_{xy}}} \right|\bm{\Phi }} \right)P\left( {\left. {{I_{yz}}} \right|\bm{\Phi }} \right)P\left( {\left. {{I_{zx}}} \right|\bm{\Phi }} \right)\\
= & \mathop {\arg \max }\limits_{\bm{\Phi }} P\left( {\left. \bm{\Phi } \right|{I_{xy}}} \right)P\left( {\left. \bm{\Phi } \right|{I_{yz}}} \right)P\left( {\left. \bm{\Phi } \right|{I_{zx}}} \right)\\
s.t.~\bm{\Phi } & \in \bm{\Omega }
\end{aligned}
\label{Eq_optimization}
\end{equation}
where ${\bm{\Phi}}$ is constrained to a low dimensional subspace ${\bm{\Omega}\subseteq\bm{\Lambda}}$ in order to resolve ambiguous joint estimations.

The posterior probabilities $P\left( {\left. {{{\bm{\phi}} _k}} \right|{I_{xy}}} \right)$, $P\left( {\left. {{{\bm{\phi}} _k}} \right|{I_{yz}}} \right)$ and $P\left( {\left. {{{\bm{\phi}} _k}} \right|{I_{zx}}} \right)$ can be estimated from heat-maps generated by CNNs. Now we present the details of multi-view 3D joint location regression. We first describe the methods of multi-view projection and learning in Section~\ref{Multiview_learning} and then describe the method of multi-view fusion in Section~\ref{Multiview_fusing}.

\subsection{Multi-view Projection and Learning}

\begin{figure}[t]
	\begin{center}
		\includegraphics[width=1.0\linewidth]{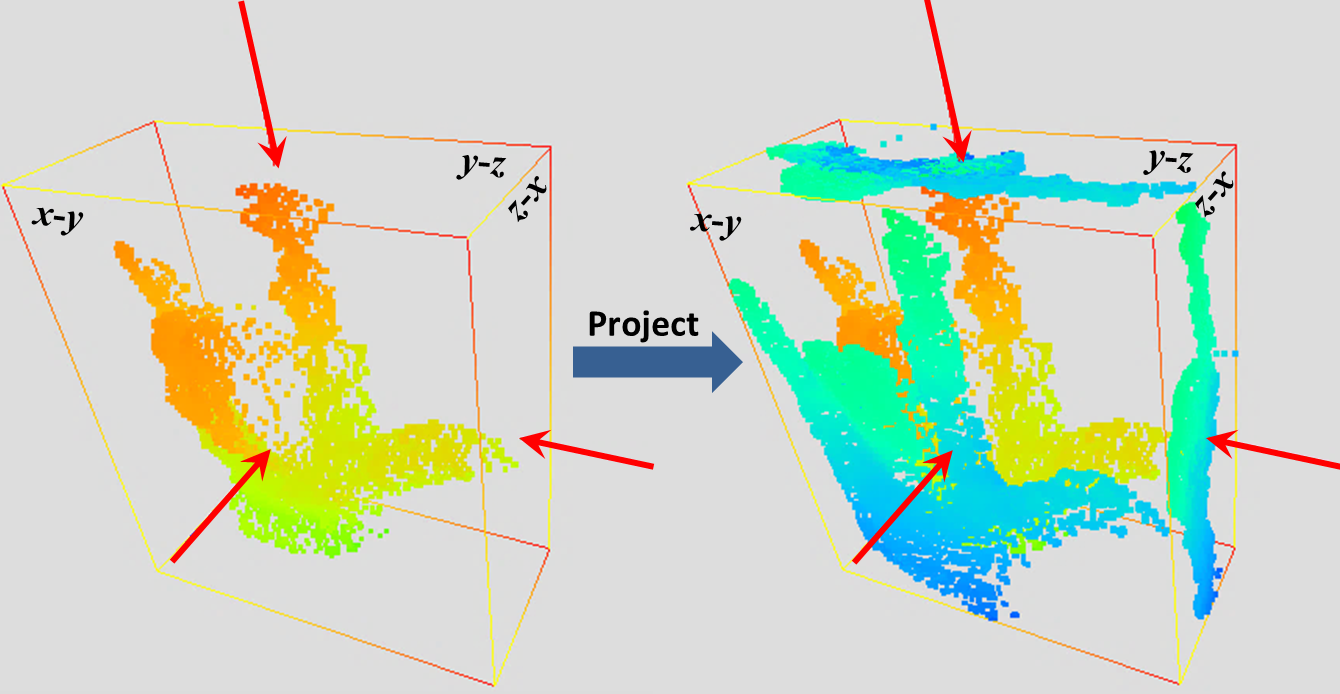}
	\end{center}
	\caption{Illustration of 3D points projection. 3D points obtained from the input depth image are projected onto \textit{x}-\textit{y}, \textit{y}-\textit{z} and \textit{z}-\textit{x} planes of the OBB coordinate system, respectively.}
	\label{fig:Fig_Projection}
\end{figure}

The objective for multi-view projection and learning is to generate projected images on each view and learn the relations between the projected images and the heat-maps of each view. First, we describe the details of 3D projections. Then, we introduce the architecture of the CNNs.

\textbf{3D Points Projection:} As illustrated in Fig.~\ref{fig:Fig_Framework}, the input depth image is first converted to a set of 3D points in the world coordinate system by using the depth camera's intrinsic parameters, e.g. the position of principal point and the focal length. To generate multi-view's projections, we project these 3D points onto three orthogonal planes. As shown in Fig.~\ref{fig:Fig_Projection}, an oriented bounding box (OBB) is generated by performing principal component analysis (PCA) on the set of 3D points, which is a tight fit around these 3D points in local space \cite{van2015essential}. The origin of OBB coordinate system is set on the center of the bounding box, and its \textit{x}, \textit{y}, \textit{z} axes are respectively aligned with the 1st, 2nd and 3rd principal components. This coordinate system is set as the projection coordinate system.

For 3D points projection onto a plane, the distances from 3D points to the projection plane are normalized between 0 and 1 (with nearest points set to 0, farthest points set to 1). Then, 3D points are orthographically projected onto the OBB coordinate system's \textit{x}-\textit{y}, \textit{y}-\textit{z} and \textit{z}-\textit{x} planes respectively, as shown in Fig.~\ref{fig:Fig_Projection}. The corresponding normalized distances are stored as pixel values of the projected images. If multiple 3D points are projected onto the same pixel, the smallest normalized distance will be stored as the pixel value. Notice that the projections on the three orthogonal planes maybe coarse because of the resolution of the depth map~\cite{Action2010Li}, which can be solved by performing median filter and opening operation on the projected images.

\begin{figure}[t]
	\begin{center}
		\includegraphics[width=1.0\linewidth]{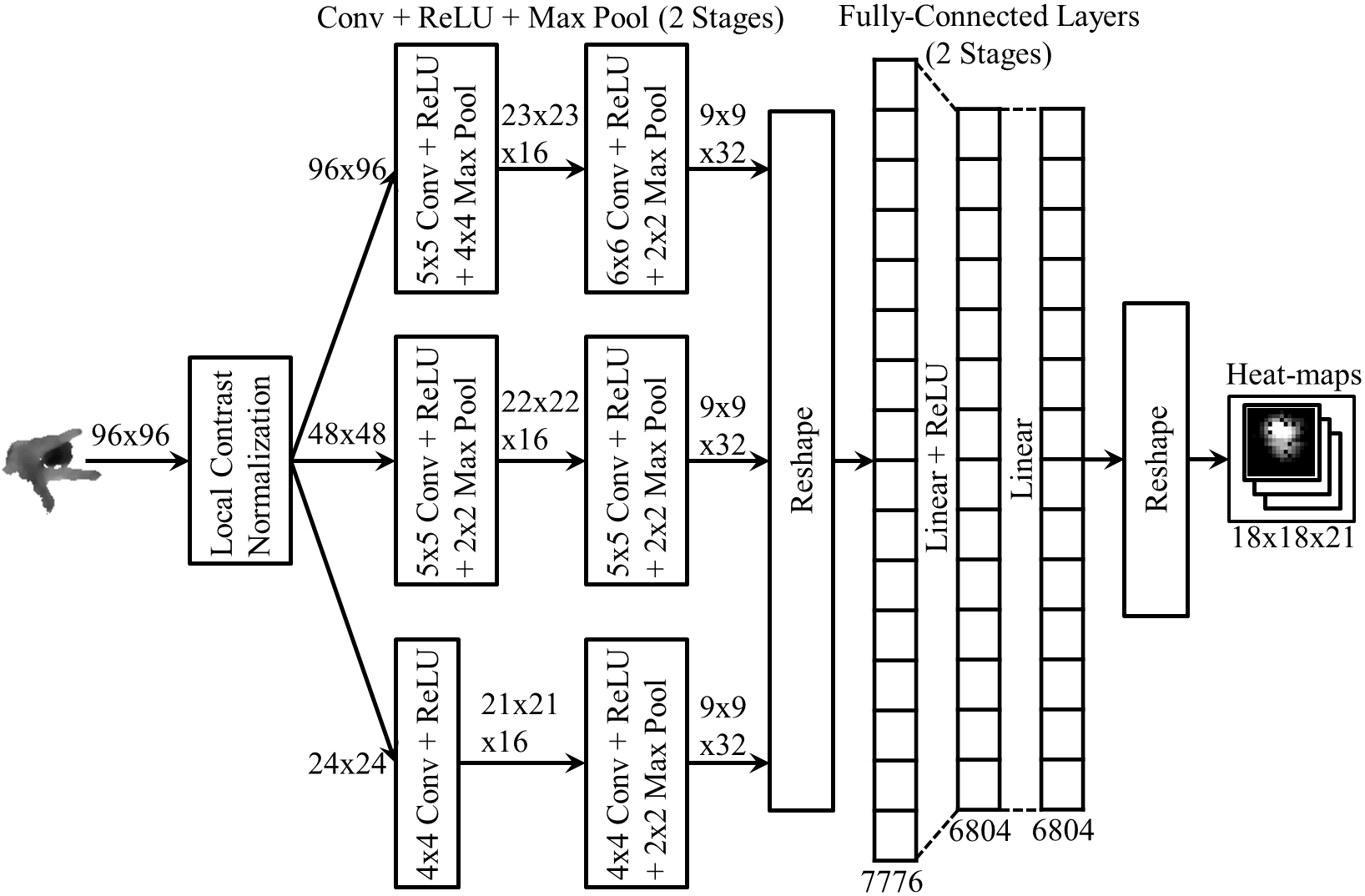}
	\end{center}
	\caption{Convolutional Network architecture for each view. The network contains convolutional layers and fully-connected layers. In convolutional layers, there are three banks for multi-resolution inputs. The network generates 21 heat-maps with the size of 18${ \times }$18 pixels. All of the three views have the same network architecture and the same architectural parameters.}
	\label{fig:Fig_CNN}
\end{figure}

\textbf{Architecture of CNNs:} Since we project 3D points onto three views, for each view, we construct a convolutional network having the same network architecture and the same architectural parameters. Inspired by the work of Tompson et al. in~\cite{tompson14tog}, we employ the multi-resolution convolutional networks architecture for each view as shown in Fig.~\ref{fig:Fig_CNN}. The input projected images are resized to 96${ \times }$96 pixels and then filtered by local contrast normalization (LCN)~\cite{jarrett2009best} to normalize the contrast in the image. After LCN, the 96${ \times }$96 image is down-sampled to 48${ \times }$48 and 24${ \times }$24 pixels. All of these three images with different resolutions are propagated through three banks which consist of two convolutional stages. The output feature maps of these three banks are concatenated and fed into a fully-connected network containing two linear stages. The final outputs of this network are 21 heat-maps with 18${ \times }$18 pixels, of which the intensity indicates the confidence of a joint locating in the 2D position on a specific view.
\label{Multiview_learning}

\subsection{Multi-view Fusion}
The objective for multi-view fusion is to estimate the 3D hand joint locations from three views' heat-maps. Let ${\phi _{kx}}$, ${\phi _{ky}}$ and ${\phi _{kz}}$ denote the \textit{x}, \textit{y} and \textit{z} coordinates of the 3D hand joint location ${{\bm{\phi}} _k}$ in the OBB coordinate system.

The CNNs generate a set of heat-maps for each \mbox{joint}, each view. Since the intensity on a heat-map indicates the confidence of a joint locating in the 2D position of the \textit{x}-\textit{y}, \textit{y}-\textit{z} or \textit{z}-\textit{x} view, we can get the corresponding probabilities $P\left( {\left. {{\phi _{kx}},{\phi _{ky}}} \right|{I_{xy}}} \right)$, $P\left( {\left. {{\phi _{ky}},{\phi _{kz}}} \right|{I_{yz}}} \right)$, and $P\left( {\left. {{\phi _{kz}},{\phi _{kx}}} \right|{I_{zx}}} \right)$ from three views' heat-maps.

Assuming that, conditioned on the \textit{x}-\textit{y} view, the distribution of \textit{z} variable is uniform, we have:
\begin{equation}
\begin{aligned}
P\left( {\left. {{{\bm{\phi}} _k}} \right|{I_{xy}}} \right) & = P\left( {\left. {{\phi _{kx}},{\phi _{ky}},{\phi _{kz}}} \right|{I_{xy}}} \right) \\
& = {P\left( {\left. {{\phi _{kx}},{\phi _{ky}}} \right|{I_{xy}}} \right)}P\left( {\left. {{\phi _{kz}}} \right|{I_{xy}}} \right)\\
& \propto P\left( {\left. {{\phi _{kx}},{\phi _{ky}}} \right|{I_{xy}}} \right)
\end{aligned}
\end{equation}
With similar assumptions, for the other two views, it can be derived that $P\left( {\left. {{{\bm{\phi}} _k}} \right|{I_{yz}}} \right) \propto P\left( {\left. {{\phi _{ky}},{\phi _{kz}}} \right|{I_{yz}}} \right)$ and $P\left( {\left. {{{\bm{\phi}} _k}} \right|{I_{zx}}} \right) \propto P\left( {\left. {{\phi _{kz}},{\phi _{kx}}} \right|{I_{zx}}} \right)$.

We assume that the hand joint locations are independent conditioned on each view's projected image. Thus, the optimization problem in Eq.~\ref{Eq_optimization} can be transformed into:
\begin{equation}
\begin{aligned}
{\bm{\Phi }^ * } & = \mathop {\arg \max }\limits_{\bm{\Phi }} P\left( {\left. \bm{\Phi } \right|{I_{xy}}} \right)P\left( {\left. \bm{\Phi } \right|{I_{yz}}} \right)P\left( {\left. \bm{\Phi } \right|{I_{zx}}} \right)\\
& = \mathop {\arg \max }\limits_{\bm{\Phi }} \prod\limits_k {P\left( {\left. {{{\bm{\phi}} _k}} \right|{I_{xy}}} \right)P\left( {\left. {{{\bm{\phi}} _k}} \right|{I_{yz}}} \right)P\left( {\left. {{{\bm{\phi}} _k}} \right|{I_{zx}}} \right)} \\
& = \mathop {\arg \max }\limits_{\bm{\Phi }} \prod\limits_k {Q\left( {{\phi _{kx}},{\phi _{ky}},{\phi _{kz}}} \right)}
\end{aligned}
\label{Eq_6}
\end{equation}
where $Q\left( {{\phi _{kx}},{\phi _{ky}},{\phi _{kz}}} \right)$ denotes the product of probabilities $P\left( {\left. {{\phi _{kx}},{\phi _{ky}}} \right|{I_{xy}}} \right)$, $P\left( {\left. {{\phi _{ky}},{\phi _{kz}}} \right|{I_{yz}}} \right)$, and $P\left( {\left. {{\phi _{kz}},{\phi _{kx}}} \right|{I_{zx}}} \right)$ for each joint.

Eq.~\ref{Eq_6} indicates that we can get the optimal hand joint locations by maximizing the product of $Q\left( {{\phi _{kx}},{\phi _{ky}},{\phi _{kz}}} \right)$ for all the joints which can be calculated from the intensities of three views' heat-maps. In this work, a set of 3D points in the bounding box is uniformly sampled and projected onto three views to get its corresponding heat-map intensities. Then the value of $Q\left( {{\phi _{kx}},{\phi _{ky}},{\phi _{kz}}} \right)$ for a 3D point can be computed.

For simplicity of this problem, the product of probabilities $Q\left( {{\phi _{kx}},{\phi _{ky}},{\phi _{kz}}} \right)$ is approximated as a 3D Gaussian distribution $\mathcal{N}({\bm{\mu }_k},{\bm{\Sigma}_k})$, where ${\bm{\mu }_k}$ is the mean vector, ${\bm{\Sigma}_k}$ is the covariance matrix. These parameters of the Gaussian distribution can be estimated from the sampled data.

Based on above assumptions and derivations, the optimization problem in Eq.~\ref{Eq_6} can be approximated as follow:
\begin{equation}
\begin{aligned}
{\bm{\Phi }^ * } & = \mathop {\arg \max }\limits_{\bm{\Phi }} \sum\limits_k {\log Q\left( {{\phi _{kx}},{\phi _{ky}},{\phi _{kz}}} \right)} \\
& = \mathop {\arg \max }\limits_{\bm{\Phi }} \sum\limits_k {\log {\mathcal{N}({\bm{\mu }_k},{\bm{\Sigma}_k})}} \\ 
& = \mathop {\arg \min }\limits_{\bm{\Phi }} \sum\limits_k {{{\left( {{{\bm{\phi}} _k} - {\bm{\mu }_k}} \right)}^T}\bm{\Sigma }_k^{ - 1}\left( {{{\bm{\phi}} _k} - {\bm{\mu }_k}} \right)} \\
{s.t.}\ & \bm{\Phi}=\sum\nolimits_{m=1}^M {{\alpha _m}{\bm{e}_m}}  + \bm{u} \\
\end{aligned}
\end{equation}
where ${\bm{\Phi }}$ is constrained to take the linear form. In order to learn the low dimensional subspace ${\bm{\Omega }}$ of hand configuration constrains, PCA is performed on joint locations in the training dataset~\cite{liang2014resolving}. ${\bm{E}={\left[ {\bm{e}_1,\bm{e}_2, \cdots ,\bm{e}_M} \right]}}$ are the principal components, ${\bm{\alpha}={\left[ {\alpha _1,\alpha _2, \cdots ,\alpha _M} \right]^T}}$ are the coefficients of the principal components, ${\bm{u}}$ is the empirical mean vector, and ${M \ll 3 \times K}$.

As proved in the supplementary material, given the linear constrains of ${\bm{\Phi }}$, the optimal coefficient vector ${{\bm{\alpha} ^*} = {\left[ {\alpha _1^*,\alpha _2^*, \cdots ,\alpha _M^*} \right]^T}}$ is:
\begin{equation}
{\bm{\alpha} ^*} = {{\bf{A}}^{ - 1}}\bm{b}
\end{equation}
where $\textbf{A}$ is a ${M \times M}$ symmetric matrix, $\bm{b}$ is an ${M}$-dimensional column vector:
\[{\textbf{A}_{ij}} = \sum\limits_k {\bm{e}_{j,k}^T\bm{\Sigma }_k^{ - 1}{\bm{e}_{i,k}}},\ {\bm{b}_i} = \sum\limits_k {{{\left( {{\bm{\mu }_k} - {\bm{u}_k}} \right)}^T}\bm{\Sigma }_k^{ - 1}{\bm{e}_{i,k}}} \]
${\bm{e}_i} = {\left[ {\bm{e}_{i,1}^T, \bm{e}_{i,2}^T, \cdots ,\bm{e}_{i,K}^T} \right]^T};\ \bm{u} = {\left[ {\bm{u}_1^T, \bm{u}_2^T, \cdots ,\bm{u}_K^T} \right]^T}$;\ ${i,\ j=1,\ 2, \cdots,\ M}$.

The optimal joint locations ${\bm{\Phi }^*}$ are reconstructed by back-projecting the optimal coefficients ${\bm{\alpha}^*}$ in the subspace $\bm{\Omega }$ to the original joint space $\bm{\Lambda }$:
\begin{equation}
{\bm{\Phi}^*}=\sum\nolimits_{m=1}^M {{\alpha _m^*}{\bm{e}_m}}  + \bm{u}
\end{equation}

To sum up, the proposed multi-view fusing method consists of two main steps. The first step is to estimate the parameters of Gaussian distribution for each joint using the three views' heat-maps. The second step is to calculate the optimal coefficients ${\bm{\alpha}^*}$ and reconstruct the optimal \mbox{joint} locations ${\bm{\Phi}^*}$. The principal components and the empirical mean vector of hand joint configuration are obtained by applying PCA on training data during the training stage.

\label{Multiview_fusing}

\section{Experiments}
\subsection{CNNs Training}
The CNNs of multiple views described in Section~\ref{Multiview_learning} were implemented within the Torch7~\cite{collobert2011torch7} framework. The optimization algorithm applied in CNNs training process is stochastic gradient descent (SGD) with a mean squared error (MSE) loss function, since the task of hand pose estimation is a typical regression problem. For training \mbox{parameters}, we choose the batch size as 64, the learning rate as 0.2, the momentum as 0.9 and the weight decay as 0.0005. Training is stopped after 50 epochs to prevent overfitting. We use a workstation with two Intel Xeon processors, 64GB of RAM and two Nvidia Tesla K20 GPUs for CNNs training. The CNNs of three views can be trained at the same time since they are in parallel. Training the CNNs takes approximately 12 hours.

\subsection{Dataset and Evaluation Metric}
We conduct a self-comparison and a comparison with state-of-the-art methods on the dataset released in~\cite{sun2015cascaded}, which is the most challenging hand pose dataset in the literature. This dataset contains 9 subjects and each subject contains 17 gestures. In the experiment, we use 8 \mbox{subjects} as the training set for CNNs training and the remaining subject as the testing set. This is repeated 9 times for all subjects.

In addition, we conduct a cross-dataset evaluation by using the training data from the dataset in~\cite{sun2015cascaded} and the testing data from another dataset in~\cite{qian2014realtime}.

We employ two metrics to evaluate the regression performance. The first metric is the mean error distance for each joint across all the test samples, which is a standard evaluation metric. The second metric is the proportion of good test samples in the entire test samples. A test sample is regarded as good only when all the estimated joint locations are within a maximum allowed distance from the ground truth, namely the error tolerance. This worst case accuracy proposed in~\cite{Taylor2012vitruvian} is very strict.

\subsection{Self-comparisons}
For self-comparison, we implement two baselines: the single view regression approach and the multi-view regression approach using a coarse fusion method. In the single view regression approach, only the projected images on OBB coordinate system's \textit{x}-\textit{y} plane are fed into the CNNs. From the output heat-maps, we can only estimate the \textit{x} and \textit{y} coordinates of joint locations by using the Gaussian fitting method proposed in~\cite{tompson14tog}. The \textit{z} coordinate can be estimated from the intensity of the projected image. If the 2D point with the estimated \textit{x}, \textit{y} coordinates is on the background of the projected image, the \textit{z} coordinate will be specified as zero in OBB coordinate system instead of the maximum depth value, which can reduce the estimation errors on \textit{z} direction. The multi-view regression approach using a coarse fusion method can be considered as a degenerated variant of our fine fusion method. This method estimates the 3D hand joint locations by simply averaging the estimated \textit{x}, \textit{y} and \textit{z} coordinates from three views' heat-maps.

\begin{figure*}
	\begin{center}
		\includegraphics[width=1.0\linewidth]{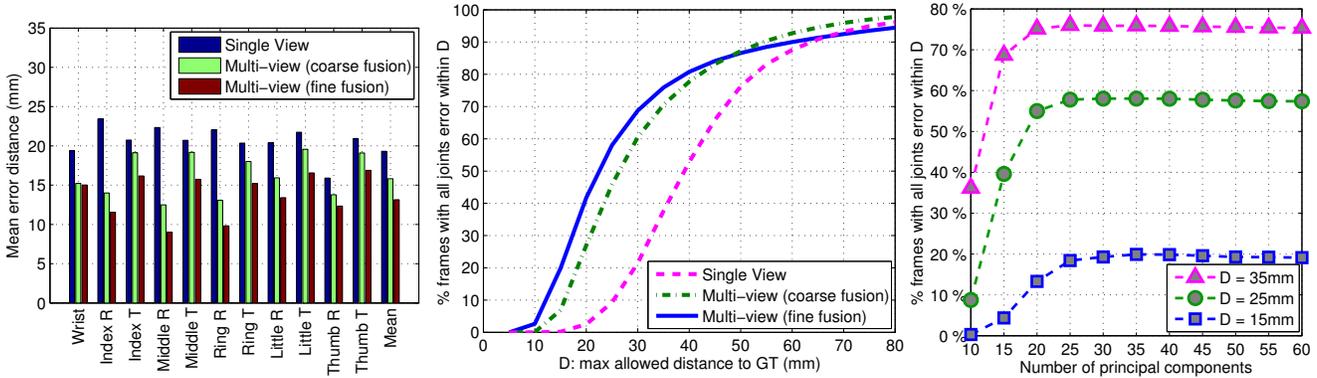}
	\end{center}
	\caption{Self-comparison of different methods on the dataset in~\cite{sun2015cascaded}. \textbf{Left}: the mean error distance for each joint across all the test samples (R:root, T:tip). \textbf{Middle}: the proportion of good test samples in the entire test samples over different error tolerances. \textbf{Right}: the impact of different number of principal components used in joint constraints on accuracy performance.}
	\label{fig:self_compare}
\end{figure*}

\begin{figure}[t]
	\begin{center}
		\includegraphics[width=1.0\linewidth]{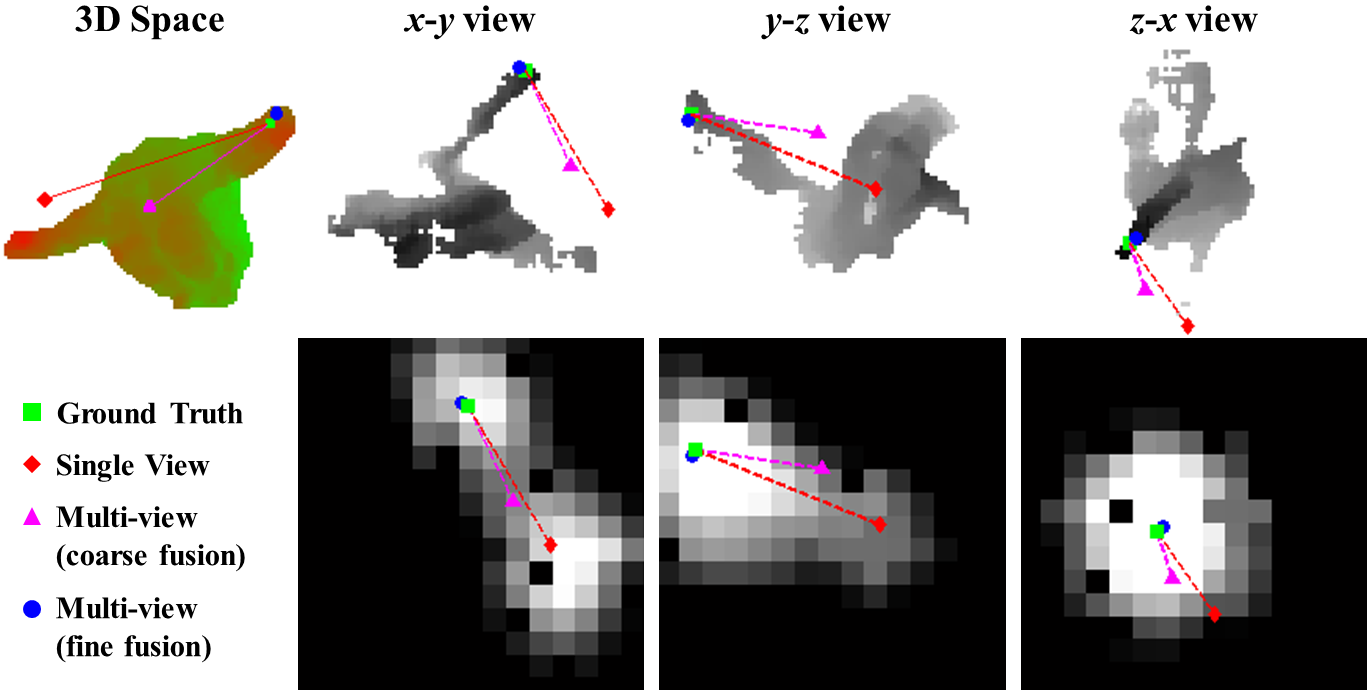}
	\end{center}
	\caption{An experimental example for self-comparison. \textbf{Top-left}: 3D point cloud with ground truth and estimated 3D locations. \textbf{Top-right}: Projection images in three views. \textbf{Bottom-right}: Heat-maps of three views. The ground truth and estimated 3D locations are back-projected onto three views and their heat-maps for comparison. Lines indicate the offsets between ground truth and estimations.}
	\label{fig:fingertip}
\end{figure}

We compare the accuracy performance of these two approaches with the multi-view fine fusion method described in Section~\ref{Methodology}. The mean error for each joint and the worst case accuracy of these three methods are shown in Fig.~\ref{fig:self_compare} (left and middle) respectively. As can be seen, the multi-view regression is effective since our two multi-view regression approaches significantly outperform the single view regression method. In addition, the fine fusion method is better than the coarse fusion method when considering the mean error performance, which is about 13 mm on the dataset in~\cite{sun2015cascaded}. When considering the worst case accuracy, the fine fusion method performs worse than the coarse fusion method only when the error tolerance is large. However, the high accuracy corresponding to small values of error tolerance should be more favorable, because the large values of error tolerance indicate that imprecise estimations will be considered as good test samples. Thus, the fine fusion method is overall better than the coarse fusion method and we apply this fusion method in the following experiments.

\begin{figure*}
	\begin{center}
		\includegraphics[width=1.0\linewidth]{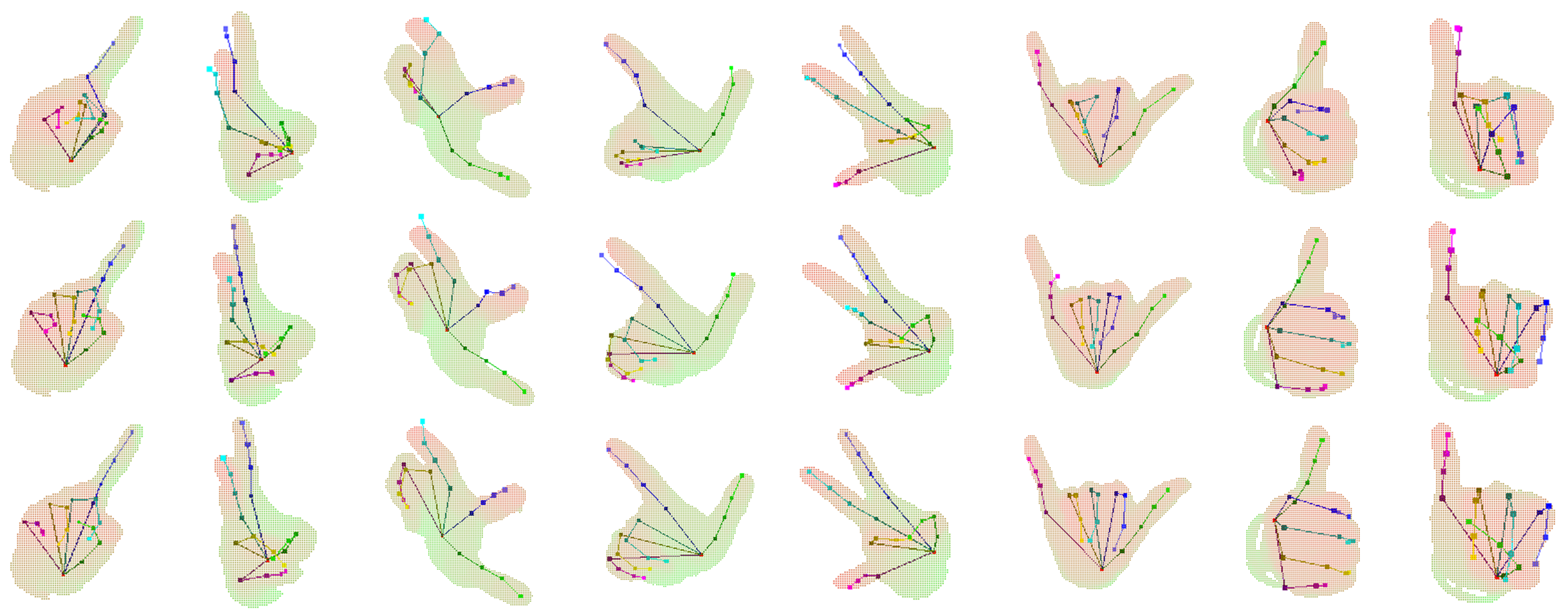}	
	\end{center}
	\caption{Qualitative results for dataset in~\cite{sun2015cascaded} of three approaches: single view regression (in the first line), our multi-view regression with coarse fusion (in the second line) and our multi-view regression with fine fusion (in the third line). We show the estimated hand joint locations on the depth image. Different hand joints and bones are visualized using different colors. This image is best viewed in color.}
	\label{fig:qualitative_results}
\end{figure*}

Fig.~\ref{fig:fingertip} shows an example of the ambiguous situation where the index fingertip is very likely to be confused with the little fingertip. As can be seen, the single view regression method only utilizes the $\textit{x}$-$\textit{y}$ view's heat-map which contains two hotspots and gives an estimation with large error distance to the ground truth. However, the multi-view fine fusion method fuses the heat-maps of three views and estimates the 3D location with high accuracy. The multi-view coarse fusion method gives an estimation in between the results of the above two methods due to its underutilization of heat-maps' information. Fig.~\ref{fig:qualitative_results} shows qualitative results of these three methods on several challenging examples to further illustrate the superiority of the multi-view fine fusion method over the other two methods.

In addition, we study the impact of different number of principal components used in joint constraints on the worst case accuracy under different error tolerances, as shown in Fig.~\ref{fig:self_compare} (right). It is reasonable to use 35 principal components in joint constraints considering the worst case accuracy. We use this setting in all the other experiments.

\subsection{Comparison with State-of-the-art}
We compare our multi-view fine fusion method with \mbox{two} state-of-the-art methods on the dataset in~\cite{sun2015cascaded}. The first method is the CNNs based hand pose estimation proposed in~\cite{tompson14tog}. The second method is the random forest based hierarchical hand pose regression proposed in~\cite{sun2015cascaded}.

\begin{figure}[t]
	\begin{center}
		\includegraphics[width=1.0\linewidth]{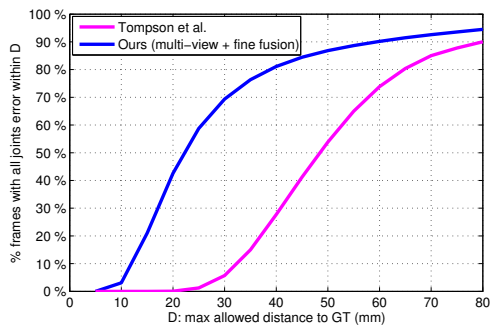}
	\end{center}
	\caption{Comparison with the approach proposed in~\cite{tompson14tog}. In this method, 14 hand joints are estimated. For fair comparison, in our method, 14 corresponding joints of 21 estimated joints are used to calculate the worst case accuracy.}
	\label{fig:other_compare_tompson}
\end{figure}

\begin{figure*}
	\begin{center}
		\includegraphics[width=1.0\linewidth]{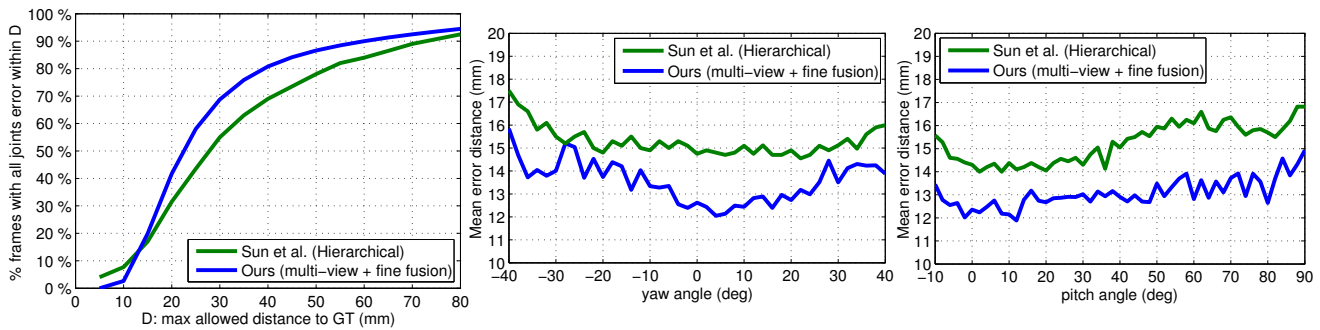}
	\end{center}
	\caption{Comparison with the approach proposed in~\cite{sun2015cascaded}. \textbf{Left}: the proportion of good test samples in the entire test samples over different error tolerances. \textbf{Middle \& right}: the mean error distance over different yaw and pitch angles of the viewpoint. Our method holds smaller average errors in all of the yaw and pitch angles. The curves of the hierarchical regression method are cropped from the results reported in~\cite{sun2015cascaded}.}
	\label{fig:other_compare}
\end{figure*}

The method in~\cite{tompson14tog} requires a model fitting process to correct large estimation errors. Since the dataset in~\cite{sun2015cascaded} does not release the hand parameters for each subject, we conduct model fitting with an uncalibrated hand model and set the hand size and finger lengths as the variables in optimization. In our implementation, this method estimates 14 hand joint locations which are a subset of the 21 hand joints used in our method. For fair comparison, we calculate the worst case accuracy of the 14 corresponding joints from the 21 joints estimated by our method. As shown in Fig.~\ref{fig:other_compare_tompson}, our multi-view regression with fine fusion method significantly outperforms the method in~\cite{tompson14tog} for the worst case accuracy. Essentially, the method in~\cite{tompson14tog} is a single view regression method which only uses the depth image as the input of the networks. This result further indicates the benefit of using multi-view's information for CNN-based 3D hand pose estimation. Even though an accurately calibrated hand model may improve the accuracy of the method in~\cite{tompson14tog} in a limited degree, it is cumbersome to calibrate the hand model for every subject and the model fitting process will increase the computational complexity.

We compare with the hierarchical regression method proposed in~\cite{sun2015cascaded}. Note that this method has been presented superior than the methods in~\cite{shotton2011realtime, tang2014latent, xu2013efficient}. Thus, we indirectly compare our method with the methods in~\cite{shotton2011realtime, tang2014latent, xu2013efficient}. As can be seen in Fig.~\ref{fig:other_compare}, our method is superior than the method in ~\cite{sun2015cascaded}. The worst case accuracy of our method is better than the method in~\cite{sun2015cascaded} over most error tolerances, as shown in Fig.~\ref{fig:other_compare} (left). Especially, when the error tolerances are 20mm and 30 mm, the good sample proportions of our method are about 10\% and 15\% higher than those of the method in~\cite{sun2015cascaded}. When the error tolerance is smaller than 15mm, the good sample proportion of our method is slightly lower than that of the method in~\cite{sun2015cascaded}. This may be caused by the relatively low resolution of the heat-maps used in our method. We also compare the average estimation errors over different viewpoint angles of these two methods. As shown in Fig.~\ref{fig:other_compare} (middle and right), the average errors of our method are smaller than those of the method in~\cite{sun2015cascaded} over all yaw and pitch viewpoint angles. In addition, our method is more robust to the pitch angle variation with a smaller standard deviation (0.64mm) than the method in~\cite{sun2015cascaded} (0.79mm).

The runtime of the entire pipeline is 14.1ms, including 2.6ms for multi-view projection, 6.8ms for CNNs forward propagation and 4.7ms for multi-view fusion. Thus, our method runs in real-time at over 70fps. Note that the process of multi-view projection and multi-view fusion is performed on CPU without parallelism, and the process of \mbox{CNNs} forward propagation is performed on GPU with parallelism for three views.

\subsection{Cross-dataset Experiment}
In order to verify the generalization ability of our CNN based multi-view regression method, we perform a cross-dataset experiment. We attempt to adapt the existing CNN based regressors learned from the source dataset in~\cite{sun2015cascaded} to a new target dataset in~\cite{qian2014realtime}.

\begin{table}[t] \small
	\begin{center}
		\begin{tabular}{cccccccc}
			\hline
			Subject        &1    &2    &3    &4    &5    &6    &Avg\\ \hline
			FORTH        &35.4     &19.8     &27.3    &26.3     &16.6     &46.2     &28.6 \\
			PSO        &29.3     &14.8     &40.2     &17.3     &16.2     &24.3     &23.6 \\
			ICP        &29.9     &20.7     &30.8     &23.9     &18.5     &32.8     &26.1 \\
			ICP-PSO    &10.1     &24.1     &13.0     &12.8     &11.9     &20.0     &15.3 \\
			ICP-PSO$^*$    &8.6     &7.4    &9.8     &10.4     &7.8    &11.7     &9.2  \\
			Ours &30.1     &19.7     &24.3     &19.9     &21.8     &20.7     &22.8 \\ \hline
		\end{tabular}
	\end{center}
	\caption{Average estimation errors (in \textit{mm}) of 6 subjects for 6 methods tested on the dataset in~\cite{qian2014realtime}.}
	\label{tab:cross_dataset}
\end{table}


In this experiment, we train the CNNs on all the 9 subjects of the dataset in~\cite{sun2015cascaded}. The CNNs are directly used for hand pose estimation on all the 6 subjects of the dataset in~\cite{qian2014realtime} by using our proposed method. According to the \mbox{evaluation} metric in~\cite{qian2014realtime}, we calculate the average errors for the wrist and the five fingertips. We compare our method with model based tracking methods reported in~\cite{qian2014realtime}, which are FORTH~\cite{oikonomidis2011efficient}, PSO~\cite{qian2014realtime}, ICP~\cite{pellegrini2008generalisation}, ICP-PSO~\cite{qian2014realtime} and ICP-PSO$^*$ (ICP-PSO with finger based initialization)~\cite{qian2014realtime}.

According to~\cite{qian2014realtime}, these model-based tracking methods need an accurate hand model that is calibrated to the size of each subject, and they rely on temporal information. Particularly, to start tracking, these methods use ground truth information to initialize the first frame. However, our method does not use such information and thus is more flexible in real scenarios and robust to tracking failure. Under such situation, our method still outperforms FORTH, PSO and ICP methods, as shown in Table~\ref{tab:cross_dataset}, which indicates that our method has good ability of generalization. It is not surprising that our method is worse than ICP-PSO and ICP-PSO$^*$, because we do not use calibrated hand model or any ground truth information or temporal information and we perform this experiment on cross-dataset which is more challenging.

\section{Conclusion}
In this paper, we presented a novel 3D hand pose regression method using multi-view CNNs. We generated a set of heat-maps of multiple views from the multi-view CNNs and fused them together to estimate 3D hand \mbox{joint} \mbox{locations}. Our multi-view approach can better leverage the 3D information in one depth image to generate accurate \mbox{estimations} of 3D locations. Experimental results showed that our method achieved superior performance for 3D hand pose estimation in real-time.


\vspace{12pt}	
\textbf{Acknowledgment:} This research, which is carried out at BeingThere Centre, is supported by Singapore National Research Foundation under its International Research Centre @ Singapore Funding Initiative and administered by the IDM Programme Office.

{\small
\bibliographystyle{ieee}
\bibliography{egbib}

\begin{thebibliography}{10}\itemsep=-1pt

\bibitem{ando2007two}
R.~K. Ando and T.~Zhang.
\newblock Two-view feature generation model for semi-supervised learning.
\newblock In {\em ICML}, 2007.

\bibitem{collobert2011torch7}
R.~Collobert, K.~Kavukcuoglu, and C.~Farabet.
\newblock Torch7: A matlab-like environment for machine learning.
\newblock In {\em BigLearn, NIPS Workshop}, 2011.

\bibitem{dollar2010cascaded}
P.~Dollár, P.~Welinder, and P.~Perona.
\newblock Cascaded pose regression.
\newblock In {\em CVPR}, 2010.

\bibitem{PTREF29}
H.~Guan, R.~S. Feris, and M.~Turk.
\newblock The isometric self-organizing map for 3d hand pose estimation.
\newblock In {\em FGR}, 2006.

\bibitem{jain2013learning}
A.~Jain, J.~Tompson, M.~Andriluka, G.~Taylor, and C.~Bregler.
\newblock Learning human pose estimation features with convolutional networks.
\newblock In {\em ICLR}, 2014.

\bibitem{jarrett2009best}
K.~Jarrett, K.~Kavukcuoglu, M.~Ranzato, and Y.~LeCun.
\newblock What is the best multi-stage architecture for object recognition?
\newblock In {\em ICCV}, 2009.

\bibitem{keskin2012hand}
C.~Keskin, F.~Kıraç, Y.~Kara, and L.~Akarun.
\newblock Hand pose estimation and hand shape classification using
  multi-layered randomized decision forests.
\newblock In {\em ECCV}, 2012.

\bibitem{PTREF38}
F.~Kirac, Y.~E. Kara, and L.~Akarun.
\newblock Hierarchically constrained 3d hand pose estimation using regression
  forests from single frame depth data.
\newblock {\em Pattern Recognition Letters}, 50:91--100, 2014.

\bibitem{Action2010Li}
W.~Li, Z.~Zhang, and Z.~Liu.
\newblock Action recognition based on a bag of 3d points.
\newblock In {\em CVPR Workshops}, 2010.

\bibitem{liang2016barehanded}
H.~Liang, J.~Wang, Q.~Sun, Y.~Liu, J.~Yuan, J.~Luo, and Y.~He.
\newblock Barehanded music: real-time hand interaction for virtual piano.
\newblock In {\em ACM SIGGRAPH I3D}, 2016.

\bibitem{liang2015parsing}
H.~Liang, J.~Yuan, and D.~Thalmann.
\newblock Parsing the hand in depth images.
\newblock {\em IEEE Transactions on Multimedia}, 16(5):1241--1253, 2014.

\bibitem{liang2014resolving}
H.~Liang, J.~Yuan, and D.~Thalmann.
\newblock Resolving ambiguous hand pose predictions by exploiting part
  correlations.
\newblock {\em IEEE Transactions on Circuits and Systems for Video Technology},
  25(7):1125--1139, 2015.

\bibitem{liang2015AR}
H.~Liang, J.~Yuan, D.~Thalmann, and N.~M. Thalmann.
\newblock Ar in hand: Egocentric palm pose tracking and gesture recognition for
  augmented reality applications.
\newblock In {\em ACM-MM}, 2015.

\bibitem{oberweger2015training}
M.~Oberweger, P.~Wohlhart, and V.~Lepetit.
\newblock Training a feedback loop for hand pose estimation.
\newblock In {\em ICCV}, 2015.

\bibitem{oikonomidis2011efficient}
I.~Oikonomidis, N.~Kyriazis, and A.~Argyros.
\newblock \mbox{Efficient} model-based 3d tracking of hand articulations using
  \mbox{Kinect}.
\newblock In {\em BMVC}, 2011.

\bibitem{pellegrini2008generalisation}
S.~Pellegrini, K.~Schindler, , and D.~Nardi.
\newblock A generalization of the icp algorithm for articulated bodies.
\newblock In {\em BMVC}, 2008.

\bibitem{qian2014realtime}
C.~Qian, X.~Sun, Y.~Wei, X.~Tang, and J.~Sun.
\newblock Realtime and robust hand tracking from depth.
\newblock In {\em CVPR}, 2014.

\bibitem{ren2013robust}
Z.~Ren, J.~Yuan, J.~Meng, and Z.~Zhang.
\newblock Robust part-based hand gesture recognition using kinect sensor.
\newblock {\em IEEE Transactions on Multimedia}, 15(5):1110--1120, 2013.

\bibitem{PTREF27}
M.~Schröder, J.~Maycock, H.~Ritter, and M.~Botsch.
\newblock Real-time hand tracking using synergistic inverse kinematics.
\newblock In {\em ICRA}, 2014.

\bibitem{shotton2011realtime}
J.~Shotton, A.~Fitzgibbon, M.~Cook, T.~Sharp, M.~Finocchio, R.~Moore,
  A.~Kipman, and A.~Blake.
\newblock Real-time human pose recognition in parts from a single depth image.
\newblock In {\em CVPR}, 2011.

\bibitem{sun2015cascaded}
X.~Sun, Y.~Wei, S.~Liang, X.~Tang, and J.~Sun.
\newblock Cascaded hand pose regression.
\newblock In {\em CVPR}, 2015.

\bibitem{supancic2015depth}
J.~S. Supancic~III, G.~Rogez, Y.~Yang, J.~Shotton, and D.~Ramanan.
\newblock Depth-based hand pose estimation: methods, data, and challenges.
\newblock In {\em ICCV}, 2015.

\bibitem{tagliasacchi2015robust}
A.~Tagliasacchi, M.~Schroeder, A.~Tkach, S.~Bouaziz, M.~Botsch, and M.~Pauly.
\newblock Robust articulated-icp for real-time hand tracking.
\newblock {\em Computer Graphics Forum}, 34(5), 2015.

\bibitem{tang2014latent}
D.~Tang, H.~J. Chang, A.~Tejani, and T.~K. Kim.
\newblock Latent regression forest: Structured estimation of 3d articulated
  hand posture.
\newblock In {\em CVPR}, 2014.

\bibitem{tang2013real}
D.~Tang, T.~H. Yu, and T.~K. Kim.
\newblock Real-time articulated hand pose estimation using semi-supervised
  transductive regression forests.
\newblock In {\em ICCV}, 2013.

\bibitem{Taylor2012vitruvian}
J.~Taylor, J.~Shotton, T.~Sharp, and A.~Fitzgibbon.
\newblock The vitruvian manifold: Inferring dense correspondences for one-shot
  human pose estimation.
\newblock In {\em CVPR}, 2012.

\bibitem{tompson2015efficient}
J.~Tompson, R.~Goroshin, A.~Jain, Y.~LeCun, and C.~Bregler.
\newblock Efficient object localization using convolutional networks.
\newblock In {\em CVPR}, 2015.

\bibitem{tompson2014joint}
J.~Tompson, A.~Jain, Y.~LeCun, and C.~Bregler.
\newblock Joint training of a convolutional network and a graphical model for
  human pose estimation.
\newblock In {\em NIPS}, 2014.

\bibitem{tompson14tog}
J.~Tompson, M.~Stein, Y.~Lecun, and K.~Perlin.
\newblock Real-time continuous pose recovery of human hands using convolutional
  networks.
\newblock {\em ACM Transactions on Graphics}, 33(5):169, 2014.

\bibitem{toshev2014deeppose}
A.~Toshev and C.~Szegedy.
\newblock Deeppose: Human pose estimation via deep neural networks.
\newblock In {\em CVPR}, 2014.

\bibitem{van2015essential}
J.~M. Van~Verth and L.~M. Bishop.
\newblock {\em Essential mathematics for games and interactive applications}.
\newblock CRC Press, 2015.

\bibitem{xu2013efficient}
C.~Xu and L.~Cheng.
\newblock Efficient hand pose estimation from a single depth image.
\newblock In {\em ICCV}, 2013.

\bibitem{xu2009multimodal}
J.~Xu, J.~Yuan, and Y.~Wu.
\newblock Multimodal partial estimates fusion.
\newblock In {\em ICCV}, 2009.

\end{thebibliography}
}

\end{document}